\documentclass[twoside,11pt]{article}
\usepackage{jair}
\usepackage{natbib}
\usepackage{hyperref}
\usepackage{adjustbox}

\usepackage{times}
\usepackage{latexsym}

\usepackage{url}
\usepackage{tabularx}

\usepackage{amsmath} 
\usepackage{caption} 
\usepackage{subcaption} 
\usepackage{multirow} 
\usepackage{amsmath} 
\usepackage{amssymb} 
\usepackage{graphicx}
\usepackage{color} 
\usepackage{arydshln} 

\usepackage{bm}
\usepackage{enumitem}
\usepackage{xspace}
\usepackage{booktabs}
\usepackage{microtype}
\newcommand{\cev}[1]{\reflectbox{\ensuremath{\vec{\reflectbox{\ensuremath{#1}}}}}}

\DeclareMathOperator*{\RNN}{\overrightarrow{\mathrm{RNN}}_{\rm src}}
\DeclareMathOperator*{\RNNL}{\overleftarrow{\mathrm{RNN}}_{\rm src}}
\DeclareMathOperator*{\RNNU}{\overrightarrow{\mathrm{RNN}}_{\rm trg}}

\newcommand{\unk}{$ \langle \textit{unk} \rangle $}
\newcommand{\bos}{$ \langle \textit{bos} \rangle $}
\newcommand{\eos}{$ \langle \textit{eos} \rangle $}
\newcommand{\pad}{$ \langle \textit{pad} \rangle $}
\newcommand{\tp}[1]{#1^\top} 
\DeclareMathOperator{\softmax}{softmax}

\newcommand{\proposed}{$ {}^{\dag} $}
\newcommand{\encdec}{EncDec\xspace}
\newcommand{\selective}{EncDec+sGate\xspace}
\newcommand{\spm}{EncDec+SPM\xspace}

\newcommand{\rencdec}{EncDec}
\newcommand{\rselective}{EncDec+sGate}
\newcommand{\rspm}{EncDec+SPM}
\newcommand{\rall}{EncDec+sGate+SPM}

\title{Source-side Prediction for Neural Headline Generation}

\newcommand{\tohoku}{$ {}^{1} $}
\newcommand{\ntt}{$ {}^{2} $}
\newcommand{\titech}{$ {}^{3} $}
\newcommand{\riken}{$ {}^{4} $}
\author{\name Shun Kiyono\tohoku \email kiyono@ecei.tohoku.ac.jp \\
  \name Sho Takase\ntt \email takase.sho@lab.ntt.co.jp \\
  \name Jun Suzuki\ntt \email suzuki.jun@lab.ntt.co.jp \\
  \name Naoaki Okazaki\titech \email okazaki@c.titech.ac.jp \\
  \name Kentaro Inui\tohoku\riken \email inui@ecei.tohoku.ac.jp \\
  \name Masaaki Nagata\ntt \email nagata.masaaki@lab.ntt.co.jp \\
  \addr  \tohoku Tohoku University\\
  \addr  \ntt NTT Communication Science Laboratories, NTT Corporation\\
  \addr  \titech Tokyo Institute of Technology\\
  \addr  \riken RIKEN\\
}

\date{}
\ShortHeadings{Source-side Prediction for Neural Headline Generation}{Source-side Prediction for Neural Headline Generation}

\begin{document}
\maketitle
\begin{abstract}
The encoder-decoder model is widely used in natural language generation tasks.
However, the model sometimes suffers from repeated redundant generation, misses important phrases, and includes irrelevant entities.
Toward solving these problems we propose a novel source-side token prediction module.
Our method jointly estimates the probability distributions over source and target vocabularies to capture a correspondence between source and target tokens.
The experiments show that the proposed model outperforms the current state-of-the-art method in the headline generation task.
Additionally, we show that our method has an ability to learn a reasonable token-wise correspondence without knowing any true alignments.
\end{abstract}

\section{Introduction}

The Encoder-Decoder model with the attention mechanism (\encdec)~\cite{DBLP:conf/nips/SutskeverVL14,cho-EtAl:2014:EMNLP2014,bahdanau:2015:ICLR,luong:2015:EMNLP} has been an epoch-making novel development that has led to great progress being made on many natural language generation tasks, such as machine translation~\citep{bahdanau:2015:ICLR}, dialog generation~\citep{shang:2015:ACL}, and headline generation~\citep{rush:2015:EMNLP}.
Today, \encdec and its variants are widely used as a strong baseline method in these tasks.

As often discussed in the community, \encdec sometimes generates sentences with repeating phrases or completely irrelevant phrases and the reason for their generation cannot be interpreted intuitively.
Moreover, \encdec also sometimes generates sentences that lack important phrases.
We refer to these observations as the problem of odd generation (\textit{odd-gen}) in \encdec.
The following table shows typical examples of \textit{odd-gen} actually generated by a typical \encdec.
\begin{figure}[ht]
 \centering
 \tabcolsep=1pt
 \scriptsize
\begin{tabular}{lp{66mm}}
 \toprule
  \multicolumn{2}{l}{\textbf{(1) Repeating Phrases}}  \\
 \midrule
  Gold: & {\tt duran duran group fashionable again}\\
  \encdec: & {\tt duran duran duran duran}\\
  \vspace{-5.0pt}\\
 \toprule
  \multicolumn{2}{l}{\textbf{(2) Lack of Important Phrases}}  \\
 \midrule
  Gold:& {\tt graf says goodbye to tennis due to injuries}\\
  \encdec:& {\tt graf retires}\\
  \vspace{-5.0pt}\\
 \toprule
  \multicolumn{2}{l}{\textbf{(3) Irrelevant Phrases}}  \\
 \midrule
  Gold:& {\tt u.s. troops take first position in serb-held bosnia}\\
  \encdec:& {\tt precede sarajevo}\\
 \end{tabular}
 \vspace{-10.0pt}
\end{figure}

This paper tackles for reducing the \textit{odd-gen} in the task of abstractive summarization.
In machine translation literature, coverage~\citep{tu:2016:ACL, mi:2016:EMNLP} and reconstruction~\citep{tu:2017:AAAI} are promising extensions of \encdec to address the problem of \textit{odd-gen}.
However, they cannot work appropriately on abstractive summarization.
This is because,
as discussed in previous studies, {\it e.g.},~\citet{nallapati:2016:CoNLL} and~\citet{suzuki:2017:EACL},
an abstractive summarization is a {\it lossy-compression} generation ({\it lossy-gen}) task
whereas a machine translation is a {\it loss-less} generation ({\it lossless-gen}) task.
Therefore, abstractive summarization does not hold the assumption of the equivalence of semantic information in source- and target-sides,
which is a fundamental assumption of the coverage and reconstruction methods.

Recently, \citet{zhou:2017:ACL} proposed incorporating an additional gate for selecting an appropriate set of words from given source sentence.
Moreover, \citet{suzuki:2017:EACL} introduced a module for estimating the upper-bound frequency of the target vocabulary given a source sentence.
These methods essentially address individual of the \textit{odd-gen} in \textit{lossy-gen} tasks.

In contrast to the previous studies, we propose a novel approach addressing all of the \textit{odd-gen} in \textit{lossy-gen} tasks.
The basic idea of our method is to incorporate an auxiliary module in addition to \encdec for modeling token-wise correspondence of the source and target,
which includes drops of source-side tokens.
We refer to our additional module as a Source-side Prediction Module (SPM).
We put the SPM on the decoder output layer to directly estimate the correspondence during the training process of \encdec.

We conduct experiments on a widely-used headline generation dataset~\citep{rush:2015:EMNLP} to evaluate the effectiveness of the proposed method.
We show that the proposed method outperforms the current state-of-the-art method on this dataset.
Additionally, we show that
our method has an ability to learn a reasonable token-wise correspondence without knowing any true alignments,
which may help reduce the \textit{odd-gen} of \encdec.

\section{Lossy-compression Generation}
\label{sec:headline}
We address the headline generation task introduced in~\citet{rush:2015:EMNLP}, which is a typical \textit{lossy-gen} task.
The source (input) is the first sentence of a news article, and the target (output) is the headline of the article.
Suppose $I$ and $J$ represent the numbers of tokens in the source and target.
An important assumption of the headline generation (\textit{lossy-gen}) task is that the relation $I > J$ always holds,
namely, the length of the target is shorter than that of the source.
This implies that we need to optimally select salient concepts included in given source sentence.
This selection indeed increases a difficulty of the headline generation for \encdec. 

Note that it is an essentially hard problem for \encdec to learn an appropriate paraphrasing of each concept in the source, which can be a main reason for generating an irrelevant headline.
In addition to this difficulty, \encdec also needs to manage the selection of concepts in the source;
e.g, discarding the excessive amount of concepts from the source would cause a headline being too short,
and utilizing the same concept multiple times in the source may lead a redundant headline.

\section{Encoder-Decoder Model with Attention Mechanism (\encdec)}
\label{sec:base_model}

\begin{figure*}[t]
    \center
    \includegraphics[width=\hsize, scale=0.8]{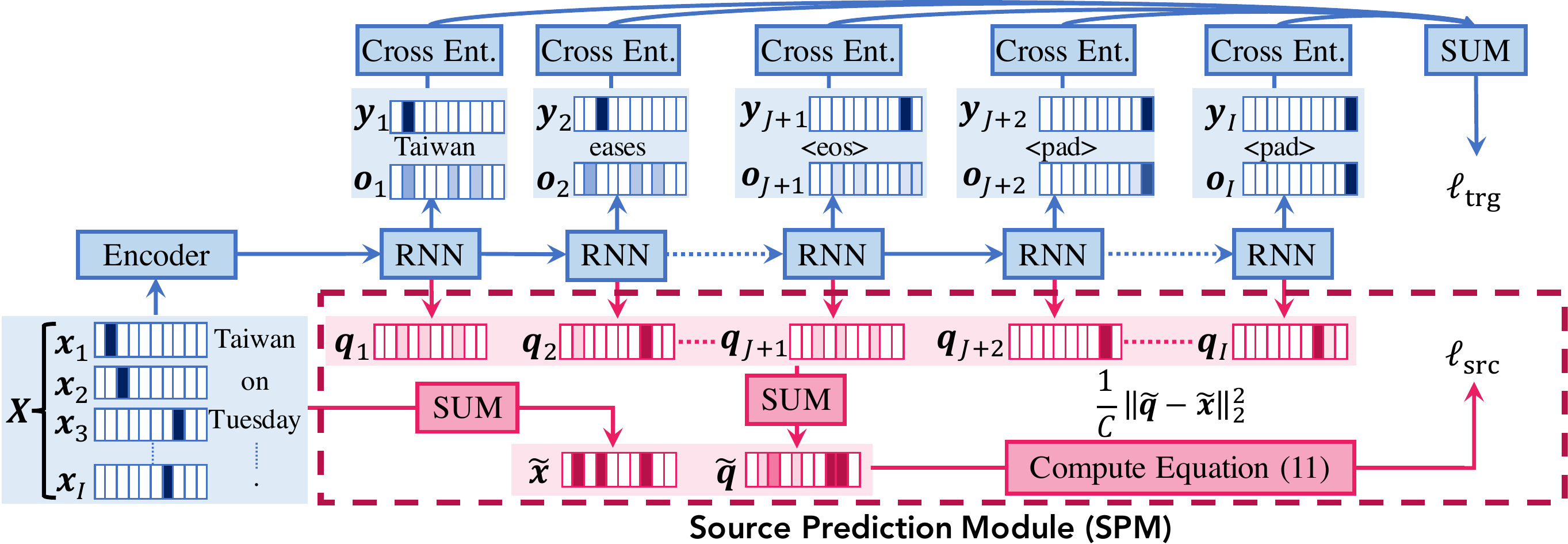}
    \caption{
    Overview of \rspm{}.
    The module inside the dashed rectangular box represents the SPM.
    The SPM predicts the probability distribution over the source vocabulary $\bm{q}_{j}$ at each time step $j$.
    After predicting all the time steps, the SPM compares the sum of the predictions $\tilde{\bm{q}}$ with the sum of the source-side tokens~$\tilde{\bm{x}}$ as an objective function $\ell_{\rm src}$.}
    \label{fig:overview}
\end{figure*}

This section briefly describes \encdec as the baseline model of our method\footnote{Our model configuration follows \encdec described in~\citet{luong:2015:EMNLP}.}.
To concisely explain \encdec, let us consider that the input of \encdec is a sequence of one-hot vectors $\bm{X}$ obtained from given source-side sentence.
Let $\bm{x}_i \in \{0,1\}^{V_{s}}$ represent the one-hot vector of $i$-th token in $\bm{X}$,
where $V_{s}$ represent a number of instances (tokens) in the source-side vocabulary $\mathcal{V}_{s}$.
We introduce $\bm{x}_{1:I}$ to represent $(\bm{x}_1,\dots,\bm{x}_{I})$ by a short notation, namely, $\bm{X} = \bm{x}_{1:I}$.
Similarly, let $\bm{y}_j \in \{0,1\}^{V_{t}}$ represent the one-hot vector of $j$-th token in the target-side sequence $\bm{Y}$,
where $V_{t}$ is a number of instances (tokens) in the target-side vocabulary $\mathcal{V}_{t}$.
Here, we define that $\bm{Y}$ always contains two additional one-hot vectors of special tokens \bos{} for $\bm{y}_0$ and \eos{} for $\bm{y}_{J+1}$, respectively.
Thus, $\bm{Y} = \bm{y}_{0:J+1}$, whose length is always $J+2$.
Then, \encdec models the following conditional probability:
\begin{align}
 p(\bm{Y} | \bm{X}) &=  \prod_{j=1}^{J+1} p( \bm{y}_{j} | \bm{y}_{0:j-1}, \bm{X})
 .
 \label{e:condpro}
\end{align}

\encdec encodes a source one-hot vector sequence $\bm{x}_{1:I}$,
and generates a hidden state sequence $\bm{h}_{1:I}$,
where $\bm{h}_{i} \in \mathbb{R}^{H}$ for all $i$, and $H$ is the size of the hidden state.
Then, the decoder with the attention mechanism computes the vector $\bm{z}_{j} \in \mathbb{R}^H$ at every decoding time step $j$ as:
\begin{align}
 \bm{z}_{j} &=  \mbox{AttnDec}(\bm{y}_{j-1}, \bm{h}_{1:I})
 .
 \label{e:atten_dec}
 \end{align}
We apply RNN cells to both encoder and the decoder.
Then, \encdec generates a target-side token based on the probability distribution $\bm{o}_j \in \mathbb{R}^{V_{t}}$ as:
\begin{equation}
  \bm{o}_j = \softmax(\bm{W}_{o}\bm{z}_{j} + \bm{b}_{o}),
  \label{e:predict}
\end{equation}
where $\bm{W}_{o} \in \mathbb{R}^{V_{t} \times H}$ is a parameter matrix and $\bm{b}_{o} \in \mathbb{R}^{V_{t}}$ is a bias term\footnote{For more detailed definitions of the encoder, decoder, and attention mechanism, see Appendices \ref{sec:append_encoder} and \ref{sec:append_decoder}, respectively.}.

To train \encdec, let $\mathcal{D}$ be a training data of headline generation that consists of source-headline sentence pairs.
Let $\theta$ represent all parameters in \encdec.
Then, we seek the optimal parameter set $\hat{\theta}$ that minimizes the following objective function $G_{1}(\theta)$ on the given training data $\mathcal{D}$:
\begin{align}
 G_{1}(\theta) &= \frac{1}{\vert \mathcal{D} \vert} \sum_{(\bm{X}, \bm{Y}) \in \mathcal{D}} \ell_{\rm trg} (\bm{Y}, \bm{X}, \theta), \nonumber\\
 \ell_{\rm trg}(\bm{Y}, \bm{X}, \theta) &= - \log \Big(
  p( \bm{Y} | \bm{X}, \theta) \Big)
 .
 \label{eq:loss1}
 \end{align}
Since $\bm{o}_{j}$ for each $j$ is a vector representation of the probabilities of $p( \hat{\bm{y}} | \bm{y}_{0:j-1}, \bm{X}, \theta)$ over the target vocabularies $\hat{\bm{y}} \in \mathcal{V}_{t}$, 
we can calculate $\ell_{\rm trg}$ as:
\begin{align}
 \ell_{\rm trg}(\bm{Y}, \bm{X}, \theta)
 \ &= -\sum_{j=1}^{J+1}{\bm{y}_{j}^{\top} \cdot \log{ \big( \bm{o}_{j} \big)}}
 .
 \label{eq:loss1_1}
 \end{align}

In the inference step, we search for the best target sequence with the trained parameters.
We use a beam search to find the target sequence that maximizes the product of the conditional probabilities as described in Equation \ref{e:condpro}.
Among several stopping criteria for the beam search \citep{huang:2017:EMNLP}, we adopt the widely used ``shrinking beam'' implemented in RNNsearch~\citep{bahdanau:2015:ICLR} \href{https://github.com/lisa-groundhog/GroundHog}{https://github.com/lisa-groundhog/GroundHog}.

\section{Proposed Method: Source Prediction Module (SPM)}
\label{sec:proposed}
In Section~\ref{sec:headline}, we assumed that the selection of concepts in the source is an essential part for the \textit{odd-gen}.
Thus, our basic idea is to extend \encdec that can manage the status of utilization of the concepts during the generation.
More precisely, instead of directly managing concepts since they are not well-defined,
we consider to model token-wise correspondence of the source and target,
including the information of source-side tokens that cannot be aligned to any target-side tokens.

Figure \ref{fig:overview} shows the overview of the proposed method, SPM.
During the training process of \encdec, the decoder estimates the probability distribution over source-side vocabulary, which is $\bm{q}_{j} \in \mathbb{R}^{V_{s}}$, in addition to that of the target-side vocabulary, $\bm{o}_{j} \in \mathbb{R}^{V_{t}}$, for every time step $j$.
Note that the decoder continues to estimate the distributions up to the source sequence length $I$ regardless of the target sequence length $J$.
Here, we introduce a special token \pad{} in the target-side vocabulary, and assume that \pad{} is repeatedly generated after finishing the generation of all target-side tokens as correct target tokens.
This means that we always assume that
the numbers of tokens in the source and target is the same,
and thus, our method allows to put one-to-one correspondence into practice in the \textit{lossy-gen} task.
In this way, \encdec can directly model token-wise correspondence of source- and target-side tokens on the decoder output layer, which includes the information of {\it unaligned} source-side tokens by aligning to \pad{}.

Unfortunately, standard headline generation datasets have no information of true one-to-one alignments between source- and target-side tokens.
Thus, we develop a novel method for train a token-wise correspondence model indirectly by an unsupervised learning manner.
Specifically, we minimize a sentence-level loss instead of a token-wise alignment loss.
We describe the details in the following sections.

\subsection{Model Definition}
In Figure~\ref{fig:overview}, the module inside the dashed line represents the SPM.
First, the SPM calculates a probability distribution over the source vocabulary $\bm{q}_{j} \in \mathbb{R}^{V_{s}}$ at each time step $j$ in the decoding process by using the following equation:
\begin{equation}
  \bm{q}_{j}=\softmax(\bm{W}_{q}\bm{z}_{j} + \bm{b}_{q}),
\end{equation}
where $\bm{W}_{q} \in \mathbb{R}^{V_{s} \times H}$ is a parameter matrix like $\bm{W}_{o}$ in Equation~\ref{e:predict},
and $\bm{b}_{q} \in \mathbb{R}^{V_{s}}$ is a bias term.
As described in Section~\ref{sec:base_model}, \encdec calculates a probability distribution over the target vocabulary $\bm{o}_j$ from $\bm{z}_{j}$.
Therefore, \encdec with the SPM jointly estimates the probability distributions over the source and target vocabularies from the same vector $\bm{z}_{j}$.

Next, we define $\bm{Y}' = \bm{y}_{0:I}$
as a concatenated sequence of $\bm{Y}$ and a sequence of one-hot vectors of the special token \pad{} with the length $I-(J+1)$,
where $\bm{y}_{J+1}$ is a one-hot vector of \eos{}, and $\bm{y}_{j}$ for each $j \in \{J+2,\dots,I\}$ is a one-hot vector of \pad{}.
Then, we also define $\bm{Y}'=\bm{Y}$ if and only if $J+1=I$.
Note that the length of $\bm{Y}'$ is always equal to or longer than that of $\bm{Y}$, that is, $|\bm{Y}'| \geq |\bm{Y}|$ since the headline generation always assumes $I > J$ as described in Section~\ref{sec:headline}.
Figure~\ref{fig:overview} also shows an actual example of $\bm{Y}'$.

Let $\tilde{\bm{x}}$ and $\tilde{\bm{q}}$ be the sums of the all one-hot vectors in source sequence $\bm{x}_{1:I}$ and the all prediction of the SPM $\bm{q}_{1:I}$, respectively, that is,
\begin{equation}
 \tilde{\bm{x}} = \sum_{i=1}^{I}{\bm{x}_i}
  ,
  \quad \mbox{and} \quad
  \tilde{\bm{q}}   = \sum_{j=1}^{I}{\bm{q}}_{j}
  .
\end{equation}
Note that $\tilde{\bm{x}}$ is a vector representation of the occurrence (or bag-of-words representation) of each source-side vocabulary appeared in the given source sequence.

Then, \encdec with the SPM models the following conditional probability:
\begin{equation}
 p(\bm{Y}', \tilde{\bm{x}} | \bm{X}) = p(  \tilde{\bm{x}} | \bm{Y}', \bm{X})p(\bm{Y}'| \bm{X})
  .
 \label{e:condpro_joint}
\end{equation}
We define $p( \bm{Y}' | \bm{X})$ as follows:
\begin{align}
 p(\bm{Y}'| \bm{X}) = \prod_{j=1}^{I} p( \bm{y}_{j} | \bm{y}_{0:j-1}, \bm{X})
 ,
 \label{e:condpro_encdec_spm}
\end{align}
which is identical to $p( \bm{Y} | \bm{X})$ in Equation~\ref{e:condpro} except substituting $I$ for $J$ to model the probabilities of \pad{} that appear from $j=I-(J+1)$ to $j=I$. 
Then, we define $p( \tilde{\bm{x}}| \bm{Y}', \bm{X})$ as follows:
\begin{equation}
 p( \tilde{\bm{x}}| \bm{Y}', \bm{X}) = \frac{1}{Z}\exp\left( \frac{-\| \tilde{\bm{q}} - \tilde{\bm{x}} \|_{2}^{2}}{C} \right)
 ,
 \label{e:condpro_spm}
\end{equation}
where $Z$ is a normalization term, and $C$ is a hyper-parameter that controls the sensitivity of the distribution. 

\subsection{Training SPM}
\label{subsec:train_spm}

Let $\gamma$ represent the parameter set of SPM.
Then, we define the loss function for SPM as follows:
\begin{align}
 \ell_{\rm src}(\tilde{\bm{x}}, \bm{X}, \bm{Y}', \gamma, \theta)  &= -\log \Big( p(\tilde{\bm{x}} | \bm{Y}', \bm{X},  \gamma, \theta) \Big)
 \nonumber
 .
\end{align}
From Equation~\ref{e:condpro_spm}, we can derive $\ell_{\rm src}$ as
\begin{align}
  \ell_{\rm src}(\tilde{\bm{x}}, \bm{X}, \bm{Y}', \gamma, \theta)  = \frac{1}{C} \| \tilde{\bm{q}} - \tilde{\bm{x}} \|_{2}^{2} + \log (Z).
\end{align}
We can discard the second term of the RHS, that is $\log (Z)$, since this is independent from $\gamma$ and $\theta$.

We jointly train the SPM and \encdec.
Therefore, we regard the sum of SPM loss ($\ell_{\rm src}$) and \encdec loss ($\ell_{\rm trg}$) as an objective loss function.
Formally, we train the SPM with \encdec by minimizing the following objective function $G_2$:
\begin{align}
 G_{2}(\theta, \gamma) =& \frac{1}{\vert \mathcal{D} \vert} \sum_{(\bm{X}, \bm{Y}) \in \mathcal{D}}
 \Bigr(\ell_{\rm trg} (\bm{Y}', \bm{X}, \theta) \nonumber \\
 &+ 
 \ell_{\rm src}(\tilde{\bm{x}}, \bm{X}, \bm{Y}', \gamma, \theta)  \Bigr)
 \label{e:obj_spm}
\end{align}
Intuitively, our learning framework can be interpreted as the multi-task learning of two different tasks, $\ell_{\rm trg}$ and $\ell_{\rm src}$.

\subsection{Inference}
It is unnecessary to compute the SPM for the purpose of evaluating decoded target sequences.
Thus, we can utilize the identical procedure of beam search used in the base \encdec briefly introduced in Section \ref{sec:base_model}.
Similarly, it is also unnecessary to produce \pad{} after generating \eos{}.
Thus, the actual computational cost of our method for the standard evaluation phase is exactly the same as the base \encdec. 

\section{Experiment}

\subsection{Dataset}
\label{subsec:dataset}
The origin of the headline generation dataset used in our experiments is identical to that used in~\citet{rush:2015:EMNLP},
namely, the dataset consists of pairs comprising the first sentence of each article and its headline from the annotated English Gigaword corpus~\citep{napoles:2012:AG}.

We slightly changed the data preparation procedure to achieve a more realistic and reasonable evaluation
since
the widely-used provided evaluation dataset already contains \unk{}, which is a replacement of all low frequency words. 
This is because the data preprocessing script provided by the authors of \citet{rush:2015:EMNLP}%
\footnote{\href{https://github.com/facebookarchive/NAMAS}{https://github.com/facebookarchive/NAMAS}.}
automatically converts low frequency words into \unk{}.
As a result, generating \unk{} can be treated as correct in evaluation
\footnote{In a personal communication with the first author of \citet{zhou:2017:ACL}, we found that their model decodes \unk{} in the same form as it appears in the test set, and \unk{} had a positive effect on the final performance of the model.}.
To penalize \unk{} in system outputs during the evaluation,
we removed \unk{} replacement procedure from the preprocessing script.
We believe this is a more realistic evaluation setting.
\citet{rush:2015:EMNLP} defined the training, validation and test split, which contain approximately 3.8M, 200K and 400K source-headline pairs, respectively.
We used the entire training split for training as in the previous studies.
We randomly sampled test data and validation data from the validation split since we found that the test split contains many noisy instances. 
Finally, our validation and test data consist of 8,000 and 10,000 source-headline pairs, respectively.
Note that they are relatively large compared with the previously used datasets, and they do not contain \unk{}. 

We also evaluated our experiments on the test data used in the previous studies. 
To the best of our knowledge, two test sets from the Gigaword are publicly available by~\citet{rush:2015:EMNLP}%
\footnote{\href{https://github.com/harvardnlp/sent-summary}{https://github.com/harvardnlp/sent-summary}}
 and~\citet{zhou:2017:ACL}%
\footnote{\href{https://res.qyzhou.me}{https://res.qyzhou.me}}.
Note that both test sets contain \unk{}.

Table \ref{table:data_summary} summarizes the characteristics of each dataset used in our experiments.

\begin{table}[t!]
\centering
\small
 \tabcolsep 0.8mm
 \begin{tabular}{lcrrl}
\toprule
 \                & use \unk{}? &       size & \#.ref & source (split)\\
 \midrule
 Training         & No       & 3,778,230  &  1      & Giga (train) \\
 Validation       & No       &     8,000  &  1      & Giga (valid)  \\
 Test (ours)      & No       &    10,000  &  1      & Giga (valid)  \\
 \midrule
 Test (Rush)      & Yes      &     1,951  &  1      & Giga (test)  \\
 Test (Zhou)      & Yes      &     2,000  &  1      & Giga (valid)  \\
\bottomrule
\end{tabular}
\caption{Characteristics of each dataset used in our experiments}
\label{table:data_summary}
\end{table}

\subsection{Evaluation Metric}
We evaluated the performance in ROUGE-1 (RG-1), ROUGE-2 (RG-2) and ROUGE-L (RG-L)\footnote{We restored sub-words to the standard token split for the evaluation.}.
We report the F1 value as given in a previous study%
\footnote{
ROUGE script option is: ``\texttt{-n2 -m -w 1.2}''
}.
We computed the ROUGE scores by using the official ROUGE script (version 1.5.5).

\subsection{Comparative Methods}
To investigate the effectiveness of the SPM, we evaluate the performance of the \encdec with the SPM.
In addition, we investigate whether the SPM improves the performance of the state-of-the-art method: \selective.
Thus, we compare the following methods on the same training setting.

\noindent\textbf{\encdec} This is the implementation of the base model explained in Section \ref{sec:base_model}.

\noindent\textbf{\selective} To reproduce the state-of-the-art method proposed by~\citet{zhou:2017:ACL}, we combined our re-implemented selective gate (sGate) with the encoder of \encdec.

\noindent\textbf{\rspm} We combined the SPM with the \encdec as explained in Section \ref{sec:proposed}.

\noindent\textbf{\rall} This is the combination of the SPM with the \selective.

\subsection{Implementation Details}
\begin{table}[t!]
\centering
 \tabcolsep 0.1mm
 \small
\begin{tabularx}{\columnwidth}{lX}
\toprule
Source Vocab. Size $V_{s}$    & \multicolumn{1}{r}{5131}                              \\
Target Vocab. Size $V_{t}$    & \multicolumn{1}{r}{5131}                              \\
Word Embedding Size D    & \multicolumn{1}{r}{200}                               \\
Hidden State Size H & \multicolumn{1}{r}{400}                               \\
\midrule
RNN Cell               & Long Short-Term Memory (LSTM) \citep{hochreiter:1997:long}                                       \\
Encoder RNN Unit       & 2-layer bidirectional-LSTM                                       \\
Decoder RNN Unit       & 2-layer LSTM with attention \citep{luong:2015:EMNLP}                           \\
\midrule
Optimizer              & Adam \citep{kingma:2015:ICLR}                                                 \\
Initial Learning Rate  & 0.001                                                 \\
Learning Rate Decay    & 0.5 for each epoch (after epoch 9)                    \\
Weight $C$ of $\ell_{\rm src}$    & 10                                     \\
Mini-batch Size        & 256 (shuffled at each epoch)                          \\
Gradient Clipping      & 5                                                     \\
Stopping Criterion     & max 15 epochs with early stopping                        \\
Regularization         & Dropout (rate 0.3)                                 \\
Beam Search            & Beam size 20 with the length normalization           \\
\bottomrule
\end{tabularx}
\caption{Configurations used in our experiment}
\label{table:optim}
\end{table}

\begin{table*}[t]
\small
 \centering
 \tabcolsep 1mm
\begin{adjustbox}{width=\textwidth}
\begin{tabular}{lccclccclccc}
\toprule
               & \multicolumn{3}{c}{\textbf{Gigaword Test (Ours)}} &  & \multicolumn{3}{c}{\textbf{Gigaword Test (Rush)}} &  & \multicolumn{3}{c}{\textbf{Gigaword Test (Zhou)}} \\
               \cmidrule{2-4}\cmidrule{6-8}\cmidrule{10-12}
                                   & RG-1      & RG-2     & RG-L     &  & RG-1   & RG-2   & RG-L   &  & RG-1   & RG-2   & RG-L   \\
               \midrule
\rencdec{}                         & 45.74     & 23.80    & 42.95    &  & 34.52  & 16.77  & 32.19  &  & 45.62  & 24.26  & 42.87   \\
\rselective{} (our impl. of SEASS) & 45.98     & 24.17    & 43.16    &  & 35.00  & 17.24  & 32.72  &  & 45.96  & 24.63  & 43.18   \\
\rspm{}\proposed{}                 & 46.18     & 24.34    & 43.35    &  & 35.17  & 17.07  & 32.75  &  & 46.21  & 24.78  & 43.27   \\
\rall{}\proposed{}                & \textbf{46.41}     & \textbf{24.58}    & \textbf{43.59}    &  & \textbf{35.79}  & \textbf{17.84}  & \textbf{33.34}  &  & \textbf{46.34}  & \textbf{24.85}  & \textbf{43.49} \\
\midrule
ABS~\citep{rush:2015:EMNLP}        & -       & -      & -      &  & 29.55  & 11.32  & 26.42  &  & 37.41  & 15.87  & 34.70  \\
SEASS~\citep{zhou:2017:ACL}        & -       & -      & -      &  & 36.15  & 17.54  & 33.63  &  & \textbf{46.86}  & \textbf{24.58}  & \textbf{43.53} \\
DRGD~\citep{li:2017:EMNLP}         & -       & -      & -      &  & 36.27  & \textbf{17.57}  & 33.62  &  & -    & -    & -  \\
WFE~\citep{suzuki:2017:EACL}       & -       & -      & -      &  & \textbf{36.30}  & 17.31  & \textbf{33.88}  &  & -    & -    & -  \\
conv-s2s~\citep{gehring:2017:arxiv}& -       & -      & -      &  & 35.88  & 17.48  & 33.29  &  & -    & -    & -   \\
\bottomrule
\end{tabular}
\end{adjustbox}
\caption{Full length ROUGE F1 evaluation results. The top row shows the results on our evaluation setting. $\dag$ is the proposed model. The bottom row shows published scores reported in previous studies\protect\footnotemark. Note that (1) SEASS consists of essentially the same architecture as our implemented \protect\selective, and (2) the top row is not directly comparable to the bottom row due to the difference of the preprocessing and the vocabulary settings (see discussions in Section~\protect\ref{subsec:results}).}
\label{table:result_summary}
\end{table*}
\footnotetext{
\citet{raffel:2017:ICML} also evaluates their model on Gigaword test set.
However, in a personal contact with the authors, we found that their evaluation setting, including the test data, critically differs from previous studies.
Thus, we do not present their published results on the table.
}

Table~\ref{table:optim} summarizes hyper-parameters and model configurations.
We selected the settings commonly-used in the previous studies, \textit{e.g.},~\cite{rush:2015:EMNLP,nallapati:2016:CoNLL,suzuki:2017:EACL}.

We constructed the vocabulary set using Byte-Pair-Encoding\footnote{\href{https://github.com/rsennrich/subword-nmt}{https://github.com/rsennrich/subword-nmt}} (BPE) \citep{sennrich:2016:ACL} to handle low frequency words, as it is now a common practice in neural machine translation.
The BPE merge operations are jointly learned from the source and the target.
We set the number of the BPE merge operations at $5,000$.
We used the same vocabulary set for both the source $\mathcal{V}_{s}$ and the target $\mathcal{V}_{t}$.
After applying the BPE, we found out that 0.1\% of the training split contained a longer target than source.
We removed such data before training.

\subsection{Results}
\label{subsec:results}
Table~\ref{table:result_summary} summarizes results on all test data.
We divide the table into two parts with a horizontal line.
The top row shows the results on our training procedure, and the bottom row shows the results reported in previous studies.
Note that the top row is not directly comparable to the bottom row due to the difference of the preprocessing and the vocabulary settings.

The top row of Table~\ref{table:result_summary} shows that \rspm{} outperformed both \rencdec{} and \rselective{}.
This result indicates that the SPM can improve the performance of \rencdec{}.
Moreover, it is noteworthy that \rall{} achieved the best performance in all metrics even though \rselective{} consists of essentially the same architecture as the current state-of-the-art model, \textit{i.e.}, SEASS.

The bottom row of Table \ref{table:result_summary} shows the results of previous methods.
They often achieved higher ROUGE scores than our models especially in Gigaword Test (Rush) and Gigaword Test (Zhou).
However, this does not immediately imply that our method is inferior to the previous methods.
This observation is basically derived by the inconsistency of vocabulary. 
In detail, our training data does not contain \unk{} because we adopted the BPE to construct vocabulary.
Thus, our models suffered from \unk{} in dataset when we conducted the evaluation on Gigaword Test (Rush) and Gigaword Test (Zhou).
Recall that, as described earlier, \selective has the same model architecture as SEASS.
Then, the similar observation can also be found in \rselective{} and SEASS.

\begin{figure}[t]
  \begin{minipage}[b]{\hsize}
    \centering
    \includegraphics[keepaspectratio, width=\hsize]{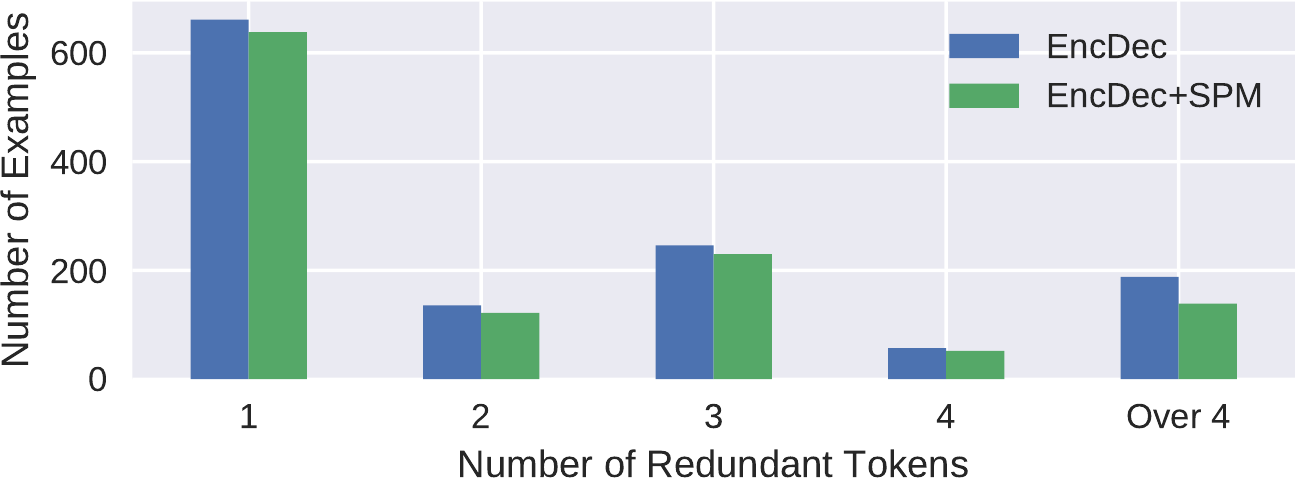}
    \subcaption{Repeating Phrases}\label{sfig:repeat_gen}
  \end{minipage}
  \begin{minipage}[b]{\hsize}
    \centering
    \includegraphics[keepaspectratio, width=\hsize]{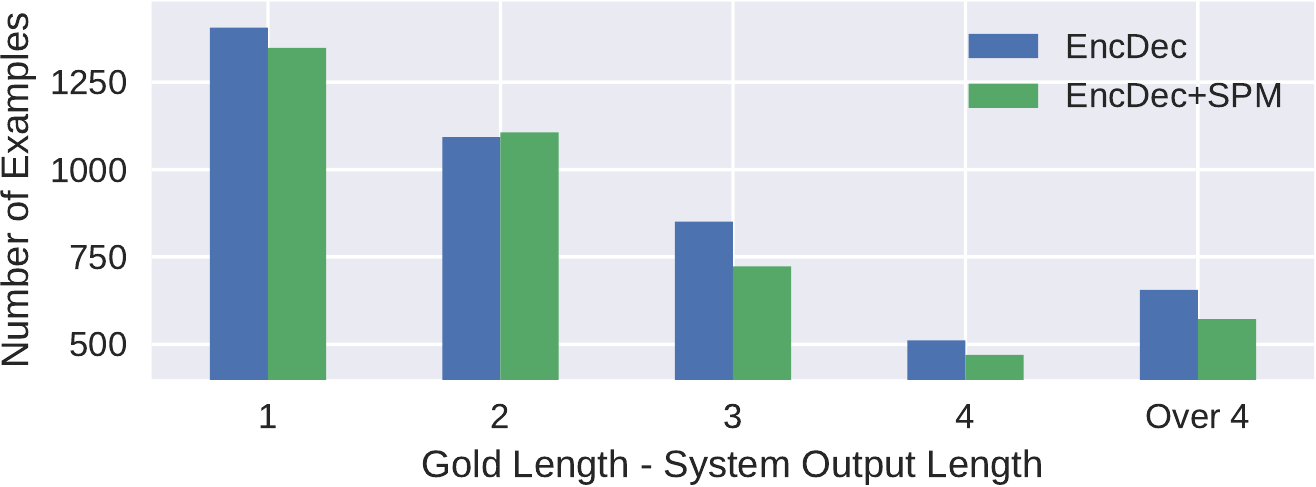}
    \subcaption{Lack of Important Phrases}\label{sfig:short_summary}
  \end{minipage}
  \caption{Comparison between \rencdec{} and \rspm{} on the number of sentences that potentially contain the \textit{odd-gen}. The smaller examples mean reduction of the \textit{odd-gen}.}\label{fig:odd_gen_reduction}
\end{figure}

\begin{figure*}[t]
 \centering
 \tabcolsep=1pt
 \scriptsize
 \begin{tabularx}{\textwidth}{lXlX}
  \midrule
  \multicolumn{4}{l}{\textbf{(1) Repeating Phrases}}  \\
  \midrule
  Gold: & \texttt{duran duran group fashionable again} & Gold: & \texttt{community college considers building \$ \#\# million technology}\\
  \rencdec{}: & \texttt{duran duran duran duran} & \rencdec{}: & \texttt{college college colleges learn to get ideas for tech center}\\
  \rspm{}: & \texttt{duran duran fashionably cool once again} & \rspm{}: & \texttt{l.a. community college officials say they 'll get ideas}\\
  \midrule
  \multicolumn{4}{l}{\textbf{(2) Lack of Important Phrases}}  \\
  \midrule
  Gold:& \texttt{graf says goodbye to tennis due to injuries} & Gold: & \texttt{new york 's primary is most suspenseful of super tuesday races}\\
  \rencdec{}:& \texttt{graf retires} & \rencdec{}: & \texttt{n.y.}\\
  \rspm{}:& \texttt{german tennis legend steffi graf retires} & \rspm{}: & \texttt{new york primary enters most suspenseful of super tuesday contests}\\
  \midrule
  \multicolumn{4}{l}{\textbf{(3) Irrelevant Phrases}}  \\
  \midrule
  Gold:& \texttt{u.s. troops take first position in serb-held bosnia} & Gold: & \texttt{northridge hopes confidence does n't wane}\\
  \rencdec{}:& \texttt{precede sarajevo} & \rencdec{}: & \texttt{csun 's csun}\\
  \rspm{}:& \texttt{u.s. troops set up first post in bosnian countryside} & \rspm{}: & \texttt{northridge tries to win northridge men 's basketball team}\\
 \end{tabularx}
 \caption{Examples of generated summary. ``Gold'' indicates the reference headline. The proposed \rspm{} model successfully reduces \textit{odd-gen}.}
 \label{fig:raw_generation}
\end{figure*}

\section{Discussion}
\label{sec:discussion}
The motivation of the SPM is to prevent the \textit{odd-gen} with one-to-one correspondence between the source and the target.
Thus, in this section, we investigate whether the SPM reduces the \textit{odd-gen} in comparison to \rencdec{}.

\subsection{Does SPM Reduce \textit{odd-gen}?}
\label{subsec:example}
For quantitative analysis, we hope to compute the statistics of generated sentences containing \textit{odd-gen}.
However, it is hard to detect the \textit{odd-gen} correctly.
Thus, we alternatively obtain a pseudo count of each type of \textit{odd-gen} as follows.

\noindent\textbf{Repeating phrases} We assume that a model causes repeating phrases if the model outputs the same token more than once.
Therefore, we compute the frequency of tokens that occur more than once in the generated headlines.
However, some phrases might occur more than once in the gold data.
To take care of this case, we subtract the frequency of tokens in the reference headline from the above calculation result.
Then, we regard the result of the subtraction as the number of repeating phrases in each generated headline.

\noindent\textbf{Lack of important phrases} We assume the generated headline which is shorter than the gold as containing the lack of important phrase.
Thus, we compute the difference of gold headline length and the generated headline length.

\noindent\textbf{Irrelevant phrases} We consider that the improvement of ROUGE scores implies the reduction of irrelevant phrases because we believe that the ROUGE penalizes irrelevant phrases.

Figure \ref{fig:odd_gen_reduction} shows the number of repeating phrases and lack of important phrases in Gigaword Test (Ours).
This figure indicates that \rspm{} reduces the \textit{odd-gen} in comparison to \rencdec{}.
Thus, we consider the SPM accomplished the reduction of the \textit{odd-gen}.
Figure \ref{fig:raw_generation} shows sampled headlines actually generated by \rencdec{} and \rspm{}.
We can clearly find that the outputs of \rencdec{} contain the \textit{odd-gen} while those of the \rspm{} do not.
These examples also demonstrate that SPM successfully reduces \textit{odd-gen}.

\subsection{Visualizing SPM and Attention}
\label{subsec:visualize}
We visualize the prediction of the SPM and the attention distribution to see the acquired token-wise correspondence between the source and the target.
Specifically, we feed the source-target pair $(\bm{X}, \bm{Y})$ to \rencdec{} and \rspm{}, and then collect the source-side prediction $(\bm{q}_{1},\dots,\bm{q}_{I})$ of \rspm{} and the attention distribution $(\bm{\alpha}_{1},\dots,\bm{\alpha}_{J})$ of \rencdec{}.
We compute the attention distribution using the following equation:
\begin{equation}
  \bm{\alpha}_{j}[i] = \frac{\exp(\tp{\bm{h}_{i}} \bm{W}_{\alpha} \vec{\bm{z}}_{j})}{\sum_{i=1}^{I}\exp(\tp{\bm{h}_{i}} \bm{W}_{\alpha} \vec{\bm{z}}_{j})}
  \label{e:context_vec}
\end{equation}
where $\bm{W}_{\alpha} \in \mathbb{R}^{H \times H}$ is a parameter matrix, and $\bm{\alpha}_{j}[i]$ denotes the $i$-th element of $\bm{\alpha}_{j}$.
Here, $\vec{\bm{z}}_{j} \in \mathbb{R}^{H}$ is the decoder hidden state.
For source-side prediction, we extracted the probability of each token $\bm{x}_{i} \in \bm{X}$ from $\bm{q}_{j}, j \in \{1,\dots,I\}$.

\begin{figure}[t]
  \begin{minipage}[b]{\hsize}
    \centering
    \includegraphics[keepaspectratio, width=\hsize]{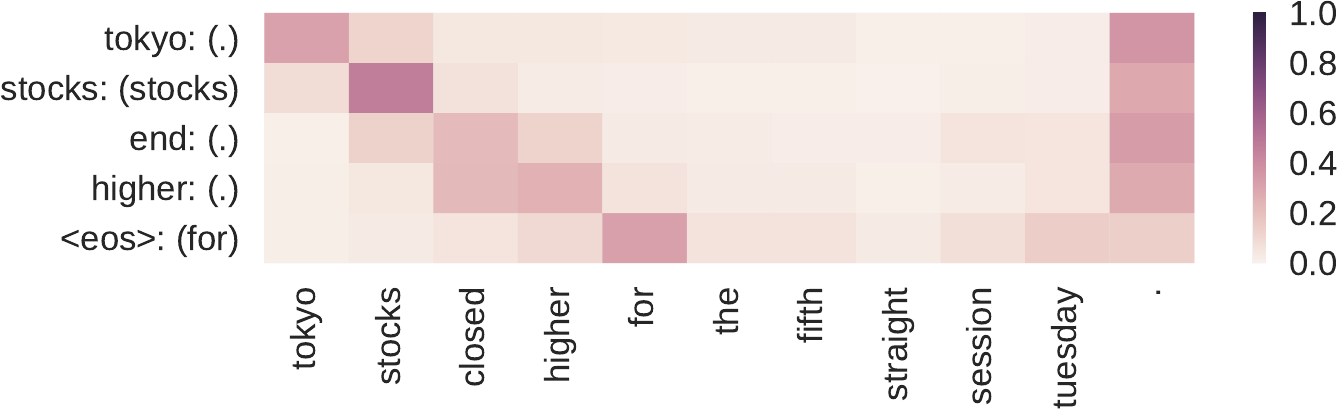}
    \subcaption{Attention distribution of \rencdec{}}\label{sfig:attn}
  \end{minipage}
  \begin{minipage}[b]{\hsize}
    \centering
    \includegraphics[keepaspectratio, width=\hsize]{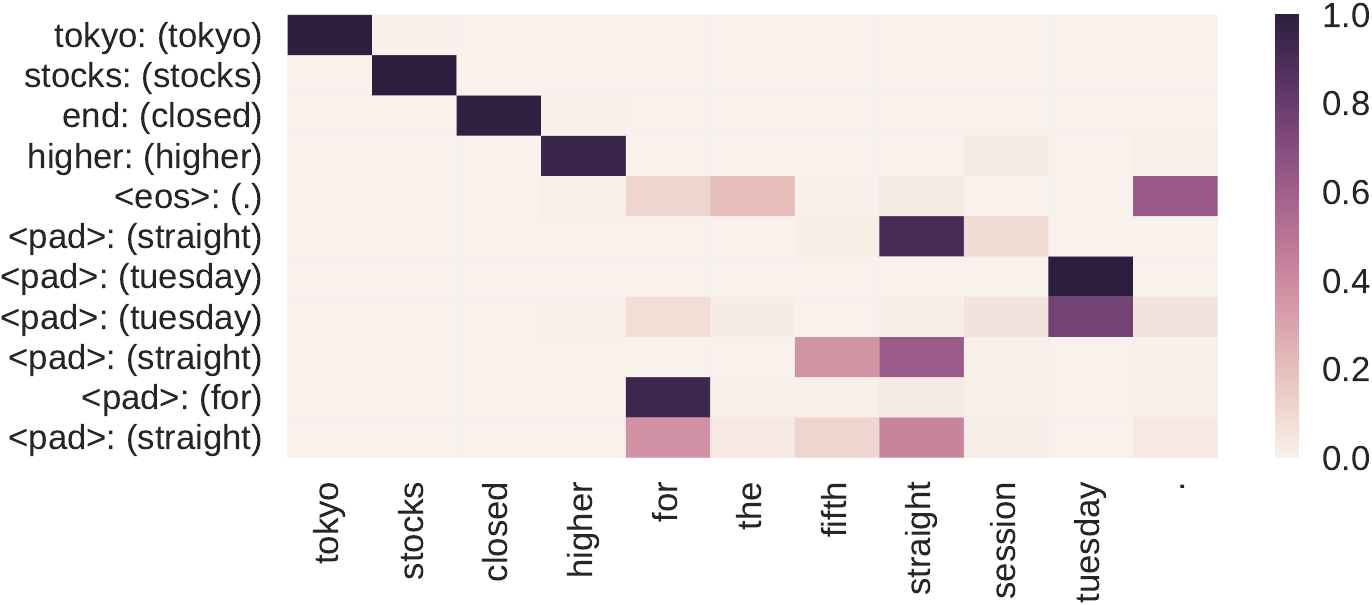}
    \subcaption{Source-side prediction of \rspm{}}\label{sfig:src}
  \end{minipage}
  \caption{
    Visualization of \rencdec{} and \rspm{}.
    The x-axis and y-axis of the figure correspond to the source sequence and the target sequence respectively.
    Token in the brackets represent the source-side token that is aligned with the target-side token of that time step.
  }\label{fig:heatmap}
\end{figure}

Figure \ref{fig:heatmap} shows an example of the heat map\footnote{For more visualizations, see Appendix \ref{sec:append_visualization}}.
We used Gigaword Test (Ours) as an input.
The brackets in the y-axis represents the source-side token that is aligned with target-side token.
We selected the aligned tokens in the following manner:
For the attention (Figure \ref{sfig:attn}), we select the token with the largest attention value.
For the SPM (Figure \ref{sfig:src}), we select the token with the largest probability over the whole vocabulary $\mathcal{V}_{s}$.

Figure \ref{sfig:attn} indicates that most of the attention distribution is concentrated at the end of the sentence.
As a result, attention provides poor token-wise correspondence between the source and the target.
For example, target-side tokens ``tokyo'' and ``end'' are both aligned with the source-side sentence period.
In contrast, Figure \ref{sfig:src} shows that the SPM provides the almost discrete correspondence between the source and the target.
The source sequence ``tokyo stocks closed higher'' is successfully aligned with the target ``tokyo stocks end higher''.
Moreover, the SPM aligned unimportant tokens for the headline such as ``straight'' and ``tuesday'' with \pad{} tokens.
Thus, this example suggests that the SPM achieved superior token-wise correspondence to the attention.
It is noteworthy that the SPM captured a one-to-one correspondence even though we trained the SPM without correct alignment information.

\section{Related Work}
In the field of neural machine translation, several methods have been proposed to solve the \textit{odd-gen}.
The coverage model~\citep{mi:2016:EMNLP,tu:2016:ACL} enforces the decoder to attend to every part of the source sequence to translate all semantic information in the source.
The reconstructor~\citep{tu:2017:AAAI} trains the translation model from the decoded target into the source.
Moreover, \citet{weng:2017:EMNLP} proposed the method to predict the untranslated words from the decoder at each time step.
These methods are designed to convert all contents in the source into a target language, since machine translation is a \textit{lossless-gen} task.
In contrast, we proposed SPM to model both paraphrasing and discarding to reduce the \textit{odd-gen} in \textit{lossy-gen} task.

We focused on the headline generation which is a well-known \textit{lossy-gen} task.
Recent studies have actively applied the \encdec to this task \citep{rush:2015:EMNLP, chopra:2016:NAACL, nallapati:2016:CoNLL}.
In the headline generation task, \citet{zhou:2017:ACL} and \citet{suzuki:2017:EACL} tackled a part of the \textit{odd-gen}.
\citet{zhou:2017:ACL} incorporated an additional gate (sGate) into the encoder to select appropriate words from the source.
\citet{suzuki:2017:EACL} proposed the frequency estimation module to reduce the repeating phrases.
Our motivation is similar to them, but we addressed solving all types of \textit{odd-gen}.
In addition, we can combine these approaches with the proposed method.
In fact, we reported in Section \ref{subsec:results} that the SPM can improve the performance of sGate with \encdec.

Apart from the \textit{odd-gen}, some studies proposed methods to improve the performance of the headline generation task.
\citet{takase:2016:EMNLP} incorporated AMR~\cite{banarescu-EtAl:2013:LAW7-ID} into the encoder to use the syntactic and semantic information of the source.
\citet{nallapati:2016:CoNLL} also encoded additional information of the source such as TF-IDF, part-of-speech tags and named entities.
\citet{li:2017:EMNLP} modeled the typical structure of a headline, such as ``Who Action What'' with a variational auto-encoder.
These approach improved the performance of the headline generation but it is unclear whether they can reduce the \textit{odd-gen}.

\section{Conclusion}
In this paper, we discussed an approach for reducing the \textit{odd-gen} in \textit{lossy-gen} tasks.
The proposed SPM learns to predict the one-to-one correspondence of tokens in the source and the target.
Experiments on the headline generation task show that the SPM improved the performance of typical \rencdec{}, and outperformed the current state-of-the-art model.
Furthermore, we demonstrated that the SPM reduced the \textit{odd-gen}.
In addition, SPM obtained token-wise correspondence between the source and the target without any alignment data.

\section*{Acknowledgments}
We thank Qingyu Zhou, Colin Raffel and Peter J. Liu for helpful discussions.
We also thank Sosuke Kobayashi for helpful comments regarding the efficient implementation.

\bibliography{jair}

\begin{thebibliography}{24}
\providecommand{\natexlab}[1]{#1}
\providecommand{\url}[1]{\texttt{#1}}
\expandafter\ifx\csname urlstyle\endcsname\relax
  \providecommand{\doi}[1]{doi: #1}\else
  \providecommand{\doi}{doi: \begingroup \urlstyle{rm}\Url}\fi

\bibitem[Sutskever et~al.(2014)Sutskever, Vinyals, and
  Le]{DBLP:conf/nips/SutskeverVL14}
Ilya Sutskever, Oriol Vinyals, and Quoc~V. Le.
\newblock {Sequence to Sequence Learning with Neural Networks}.
\newblock In \emph{Advances in Neural Information Processing Systems 27 (NIPS
  2014)}, pages 3104--3112, 2014.

\bibitem[Cho et~al.(2014)Cho, van Merrienboer, Gulcehre, Bahdanau, Bougares,
  Schwenk, and Bengio]{cho-EtAl:2014:EMNLP2014}
Kyunghyun Cho, Bart van Merrienboer, Caglar Gulcehre, Dzmitry Bahdanau, Fethi
  Bougares, Holger Schwenk, and Yoshua Bengio.
\newblock {Learning Phrase Representations using RNN Encoder--Decoder for
  Statistical Machine Translation}.
\newblock In \emph{Proceedings of the 2014 Conference on Empirical Methods in
  Natural Language Processing (EMNLP 2014)}, pages 1724--1734, 2014.

\bibitem[Bahdanau et~al.(2015)Bahdanau, Cho, and Bengio]{bahdanau:2015:ICLR}
Dzmitry Bahdanau, Kyunghyun Cho, and Yoshua Bengio.
\newblock {Neural Machine Translation by Jointly Learning to Align and
  Translate}.
\newblock In \emph{Proceedings of the 3rd International Conference on Learning
  Representations (ICLR 2015)}, 2015.

\bibitem[Luong et~al.(2015)Luong, Pham, and Manning]{luong:2015:EMNLP}
Thang Luong, Hieu Pham, and Christopher~D. Manning.
\newblock {Effective Approaches to Attention-based Neural Machine Translation}.
\newblock In \emph{Proceedings of the 2015 Conference on Empirical Methods in
  Natural Language Processing (EMNLP 2015)}, pages 1412--1421, 2015.

\bibitem[Shang et~al.(2015)Shang, Lu, and Li]{shang:2015:ACL}
Lifeng Shang, Zhengdong Lu, and Hang Li.
\newblock {Neural Responding Machine for Short-Text Conversation}.
\newblock In \emph{Proceedings of the 53rd Annual Meeting of the Association
  for Computational Linguistics and the 7th International Joint Conference on
  Natural Language Processing (ACL \& IJCNLP 2015)}, pages 1577--1586, July
  2015.

\bibitem[Rush et~al.(2015)Rush, Chopra, and Weston]{rush:2015:EMNLP}
Alexander~M. Rush, Sumit Chopra, and Jason Weston.
\newblock {A Neural Attention Model for Abstractive Sentence Summarization}.
\newblock In \emph{Proceedings of the 2015 Conference on Empirical Methods in
  Natural Language Processing (EMNLP 2015)}, pages 379--389, 2015.

\bibitem[Tu et~al.(2016)Tu, Lu, Liu, Liu, and Li]{tu:2016:ACL}
Zhaopeng Tu, Zhengdong Lu, Yang Liu, Xiaohua Liu, and Hang Li.
\newblock {Modeling Coverage for Neural Machine Translation}.
\newblock In \emph{Proceedings of the 54th Annual Meeting of the Association
  for Computational Linguistics (ACL 2016)}, pages 76--85, 2016.

\bibitem[Mi et~al.(2016)Mi, Sankaran, Wang, and Ittycheriah]{mi:2016:EMNLP}
Haitao Mi, Baskaran Sankaran, Zhiguo Wang, and Abe Ittycheriah.
\newblock {Coverage Embedding Models for Neural Machine Translation}.
\newblock In \emph{Proceedings of the 2016 Conference on Empirical Methods in
  Natural Language Processing (EMNLP 2016)}, pages 955--960, 2016.

\bibitem[Tu et~al.(2017)Tu, Liu, Shang, Liu, and Li]{tu:2017:AAAI}
Zhaopeng Tu, Yang Liu, Lifeng Shang, Xiaohua Liu, and Hang Li.
\newblock {Neural Machine Translation with Reconstruction}.
\newblock In \emph{Thirty-First AAAI Conference on Artificial Intelligence
  (AAAI 2017)}, pages 3097--3103, 2017.

\bibitem[Nallapati et~al.(2016)Nallapati, Zhou, dos Santos, Gulcehre, and
  Xiang]{nallapati:2016:CoNLL}
Ramesh Nallapati, Bowen Zhou, Cicero dos Santos, Caglar Gulcehre, and Bing
  Xiang.
\newblock {Abstractive Text Summarization Using Sequence-to-Sequence RNNs and
  Beyond}.
\newblock In \emph{Proceedings of The 20th SIGNLL Conference on Computational
  Natural Language Learning}, pages 280--290, 2016.

\bibitem[Suzuki and Nagata(2017)]{suzuki:2017:EACL}
Jun Suzuki and Masaaki Nagata.
\newblock {Cutting-off Redundant Repeating Generations for Neural Abstractive
  Summarization}.
\newblock In \emph{Proceedings of the 15th Conference of the European Chapter
  of the Association for Computational Linguistics (EACL 2017)}, pages
  291--297, 2017.

\bibitem[Zhou et~al.(2017)Zhou, Yang, Wei, and Zhou]{zhou:2017:ACL}
Qingyu Zhou, Nan Yang, Furu Wei, and Ming Zhou.
\newblock {Selective Encoding for Abstractive Sentence Summarization}.
\newblock In \emph{Proceedings of the 55th Annual Meeting of the Association
  for Computational Linguistics (ACL 2017)}, pages 1095--1104, 2017.

\bibitem[Huang et~al.(2017)Huang, Zhao, and Ma]{huang:2017:EMNLP}
Liang Huang, Kai Zhao, and Mingbo Ma.
\newblock {When to Finish? Optimal Beam Search for Neural Text Generation
  (modulo beam size)}.
\newblock In \emph{Proceedings of the 2017 Conference on Empirical Methods in
  Natural Language Processing (EMNLP 2017)}, pages 2124--2129, 2017.

\bibitem[Napoles et~al.(2012)Napoles, Gormley, and Van~Durme]{napoles:2012:AG}
Courtney Napoles, Matthew Gormley, and Benjamin Van~Durme.
\newblock {Annotated Gigaword}.
\newblock In \emph{Proceedings of the Joint Workshop on Automatic Knowledge
  Base Construction and Web-scale Knowledge Extraction}, AKBC-WEKEX '12, pages
  95--100, 2012.

\bibitem[Hochreiter and Schmidhuber(1997)]{hochreiter:1997:long}
Sepp Hochreiter and J{\"u}rgen Schmidhuber.
\newblock {Long Short-Term Memory}.
\newblock \emph{Neural Computation}, 9\penalty0 (8):\penalty0 1735--1780, 1997.

\bibitem[Kingma and Ba(2015)]{kingma:2015:ICLR}
Diederik Kingma and Jimmy Ba.
\newblock {Adam: A Method for Stochastic Optimization}.
\newblock In \emph{Proceedings of the 3rd International Conference on Learning
  Representations (ICLR 2015)}, 2015.

\bibitem[Li et~al.(2017)Li, Lam, Bing, and Wang]{li:2017:EMNLP}
Piji Li, Wai Lam, Lidong Bing, and Zihao Wang.
\newblock {Deep Recurrent Generative Decoder for Abstractive Text
  Summarization}.
\newblock In \emph{Proceedings of the 2017 Conference on Empirical Methods in
  Natural Language Processing (EMNLP 2017)}, pages 2081--2090, 2017.

\bibitem[Gehring et~al.(2017)Gehring, Auli, Grangier, Yarats, and
  Dauphin]{gehring:2017:arxiv}
Jonas Gehring, Michael Auli, David Grangier, Denis Yarats, and Yann~N Dauphin.
\newblock {Convolutional Sequence to Sequence Learning}.
\newblock \emph{arXiv preprint arXiv:1705.03122}, 2017.

\bibitem[Raffel et~al.(2017)Raffel, Luong, Liu, Weiss, and
  Eck]{raffel:2017:ICML}
Colin Raffel, Minh{-}Thang Luong, Peter~J. Liu, Ron~J. Weiss, and Douglas Eck.
\newblock {Online and Linear-Time Attention by Enforcing Monotonic Alignments}.
\newblock In \emph{Proceedings of the 34th International Conference on Machine
  Learning (ICML 2017)}, pages 2837--2846, 2017.

\bibitem[Sennrich et~al.(2016)Sennrich, Haddow, and Birch]{sennrich:2016:ACL}
Rico Sennrich, Barry Haddow, and Alexandra Birch.
\newblock {Neural Machine Translation of Rare Words with Subword Units}.
\newblock In \emph{Proceedings of the 54th Annual Meeting of the Association
  for Computational Linguistics (ACL 2016)}, pages 1715--1725, 2016.

\bibitem[Weng et~al.(2017)Weng, Huang, Zheng, Dai, and Chen]{weng:2017:EMNLP}
Rongxiang Weng, Shujian Huang, Zaixiang Zheng, Xinyu Dai, and Jiajun Chen.
\newblock {Neural Machine Translation with Word Predictions}.
\newblock In \emph{Proceedings of the 2017 Conference on Empirical Methods in
  Natural Language Processing (EMNLP 2017)}, pages 136--145, 2017.

\bibitem[Chopra et~al.(2016)Chopra, Auli, and Rush]{chopra:2016:NAACL}
Sumit Chopra, Michael Auli, and Alexander~M. Rush.
\newblock {Abstractive Sentence Summarization with Attentive Recurrent Neural
  Networks}.
\newblock In \emph{Proceedings of the 2016 Conference of the North American
  Chapter of the Association for Computational Linguistics (NAACL 2016)}, pages
  93--98, 2016.

\bibitem[Takase et~al.(2016)Takase, Suzuki, Okazaki, Hirao, and
  Nagata]{takase:2016:EMNLP}
Sho Takase, Jun Suzuki, Naoaki Okazaki, Tsutomu Hirao, and Masaaki Nagata.
\newblock {Neural Headline Generation on Abstract Meaning Representation}.
\newblock In \emph{Proceedings of the 2016 Conference on Empirical Methods in
  Natural Language Processing (EMNLP 2016)}, pages 1054--1059, 2016.

\bibitem[Banarescu et~al.(2013)Banarescu, Bonial, Cai, Georgescu, Griffitt,
  Hermjakob, Knight, Koehn, Palmer, and Schneider]{banarescu-EtAl:2013:LAW7-ID}
Laura Banarescu, Claire Bonial, Shu Cai, Madalina Georgescu, Kira Griffitt, Ulf
  Hermjakob, Kevin Knight, Philipp Koehn, Martha Palmer, and Nathan Schneider.
\newblock {Abstract Meaning Representation for Sembanking}.
\newblock In \emph{Proceedings of the 7th Linguistic Annotation Workshop and
  Interoperability with Discourse}, pages 178--186, 2013.

\end{thebibliography}

\clearpage
\appendix

\section{Baseline Model Encoder}
\label{sec:append_encoder}
We employ bidirectional RNN (BiRNN) as the encoder of the baseline model.
BiRNN is composed of two separate RNNs for forward ($\RNN$) and backward ($\RNNL$) directions.
The forward RNN reads the source sequence $\bm{X}$ from left to right order and constructs hidden states $(\vec{\bm{h}}_{1},\ldots,\vec{\bm{h}}_{I})$.
Similarly, the backward RNN reads input in the reverse order to obtain another sequence of hidden states $(\cev{\bm{h}}_{1},\ldots,\cev{\bm{h}}_{I})$.
Lastly, we take a summation of hidden states of each direction to construct final representation of the source sequence $(\bm{h}_{1},\ldots,\bm{h}_{I})$.

Concretely, for given time step $i$, the representation $\bm{h}_i$ is constructed as follows:
\begin{align}
  \vec{\bm{h}}_{i} & = \RNN (\bm{E}_{s}\bm{x}_{i}, \vec{\bm{h}}_{i-1}) \label{eq:forward_def2} ,\\
  \cev{\bm{h}}_{i} & = \RNNL (\bm{E}_{s}\bm{x}_{i}, \cev{\bm{h}}_{i+1}) \label{eq:backward_def} ,\\
  \bm{h}_{i} & = \vec{\bm{h}}_{i} + \cev{\bm{h}}_{i}
\end{align}
where $\bm{E}_{s} \in \mathbb{R}^{D\times V_{s}}$ denotes the word embedding matrix of the source-side, and $D$ denotes the size of word embedding.

\section{Baseline Model Decoder}
\label{sec:append_decoder}
The baseline model $\mbox{AttnDec}$ is composed of the decoder and the attention mechanism.
Here, the decoder is the unidirectional RNN with the input-feeding approach \citep{luong:2015:EMNLP}.
Concretely, the decoder RNN takes output from previous time step $\bm{y}_{j-1}$, decoder hidden state $\vec{\bm{z}}_{j-1}$ and final hidden state $\bm{z}_{j-1}$ to
derive a hidden state of current time step $\bm{z}_{j}$:
\begin{align}
  \vec{\bm{z}}_{j} &= \RNNU(\bm{E}_{t}\bm{y}_{j-1}, \bm{z}_{j-1}, \vec{\bm{z}}_{j-1}), \\
  \vec{\bm{z}}_{0} &= \vec{\bm{h}}_{I} + \cev{\bm{h}}_{1}
  \label{e:hs1}
\end{align}
where $\bm{E}_{t} \in \mathbb{R}^{D \times V_{t}}$ denotes the word embedding matrix of the decoder.
Here, $\bm{z}_{0}$ is defined as a zero vector.

\section{Baseline Model Attention Mechanism}
\label{sec:append_attention}
The attention architecture of the baseline model is same as the \textit{Global Attention} model proposed by \citet{luong:2015:EMNLP}.
The Attention is responsible for constructing the final hidden state $\bm{z}_{j}$ from the decoder hidden state $\vec{\bm{z}}_{j}$ and encoder hidden states $(\bm{h}_{1},\dots,\bm{h}_{I})$.

Firstly, the model computes the attention distribution $\bm{\alpha}_{j} \in \mathbb{R}^{I}$ from the decoder hidden state $\vec{\bm{z}}_{j}$ and encoder hidden states $(\bm{h}_{1},\dots,\bm{h}_{I})$.
Among three attention scoring functions proposed in \citet{luong:2015:EMNLP}, we employ \textit{general} function.
This function calculates the attention score in the bilinear form.
Specifically, the attention score between the $i$-th source hidden state and $j$-th decoder hidden state is computed by the following equation:
\begin{equation}
  \bm{\alpha}_{j}[i] = \frac{\exp(\tp{\bm{h}_{i}} \bm{W}_{\alpha} \vec{\bm{z}}_{j})}{\sum_{i=1}^{I}\exp(\tp{\bm{h}_{i}} \bm{W}_{\alpha} \vec{\bm{z}}_{j})}
  \label{e:context_vec_appendix}
\end{equation}
where $\bm{W}_{\alpha} \in \mathbb{R}^{H \times H}$ is a parameter matrix, and $\bm{\alpha}_{j}[i]$ denotes $i$-th element of $\bm{\alpha}_{j}$.

$\bm{\alpha}_{j}$ is then used for collecting the source-side information that is relevant for predicting the target token.
This is done by taking the weighted sum on the encoder hidden states:
\begin{equation}
  \bm{c}_{j} = \sum_{i=1}^{I} \bm{\alpha}_{j}[i]\bm{h}_{i}
\end{equation}

Finally, the source-side information is mixed with the decoder hidden state to derive final hidden state $\bm{z}_{j}$.
Concretely, the context vector $\bm{c}_{j}$ is concatenated with $\vec{\bm{z}}_j$ to form vector $\bm{u}_{j} \in \mathbb{R}^{2H}$.
$\bm{u}_{j}$ is then fed into a single fully-connected layer with $\tanh$ nonlinearity:
\begin{equation}
  \bm{z}_{j} = \tanh(\bm{W}_{s} \bm{u}_{j})
  \label{e:hs2}
\end{equation}
where $\bm{W}_{s} \in \mathbb{R}^{H \times 2H}$ is a parameter matrix.

\section{Extra Visualizations of SPM and Attention}
\label{sec:append_visualization}
\begin{figure*}[t]
  \begin{minipage}[b]{\hsize}
    \centering
    \includegraphics[keepaspectratio, width=\hsize]{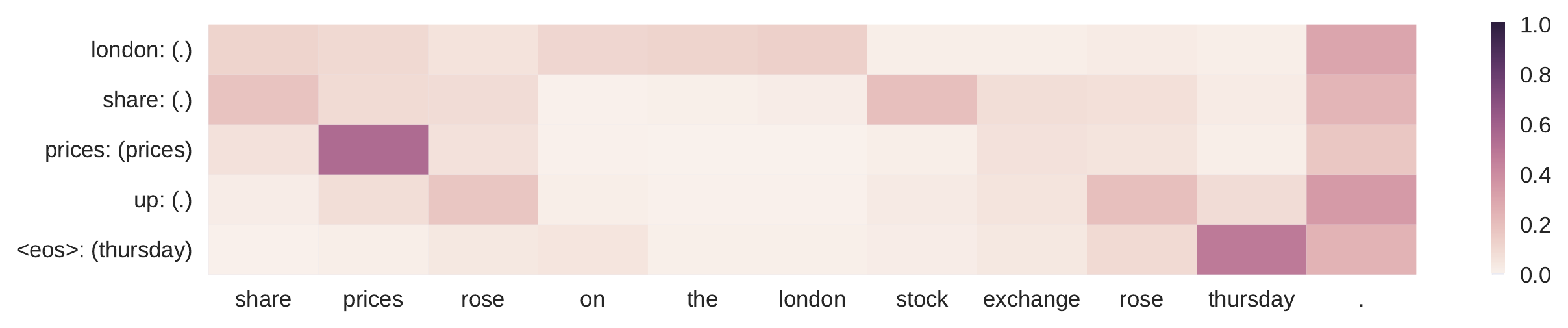}
    \subcaption{Attention distribution of \encdec}
  \end{minipage}
  \begin{minipage}[b]{\hsize}
    \centering
    \includegraphics[keepaspectratio, width=\hsize]{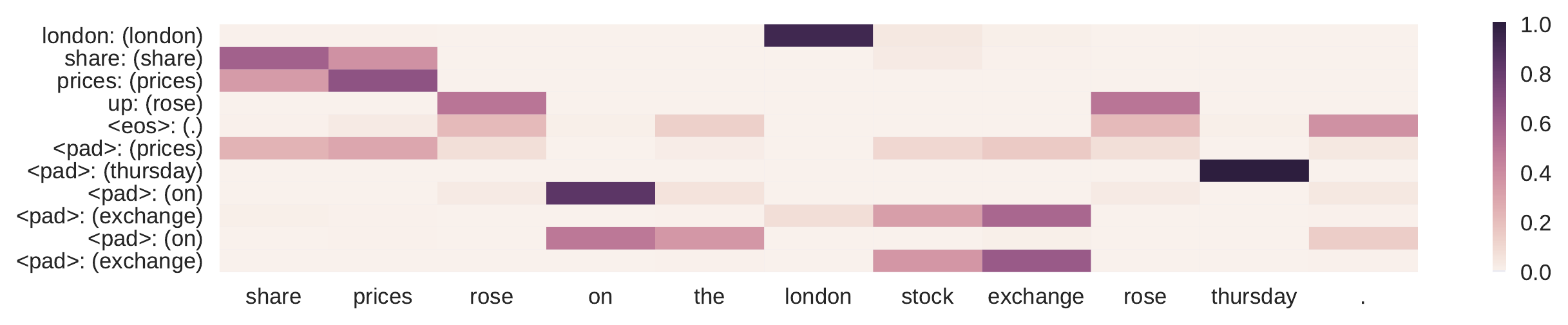}
    \subcaption{Source-side prediction of \spm}
  \end{minipage}
  \caption{
  Although ``london'' is not in the beginning of the source sentence, the SPM aligns ``london'' in the source and the target.
  On the other hand, \rencdec{} concentrates the most of the attention in the end of the sentence.
  As a result, the most of the target-side tokens are aligned with the sentence period of the source sentence.
  }\label{fig:heatmap01}
\end{figure*}

\begin{figure*}[t]
  \begin{minipage}[b]{\hsize}
    \centering
    \includegraphics[keepaspectratio, width=\hsize]{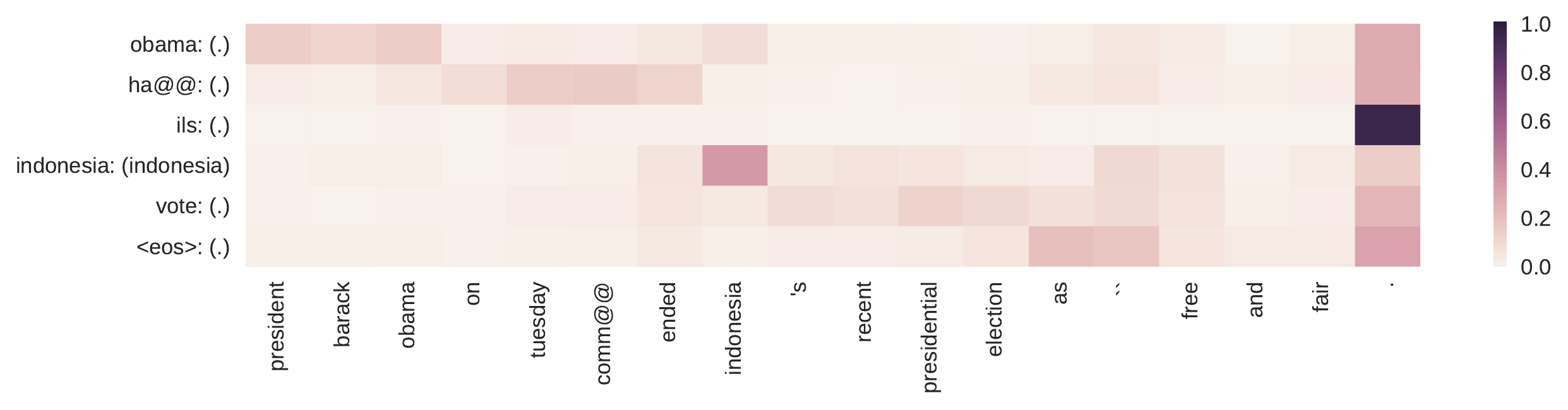}
    \subcaption{Attention distribution of \encdec}
  \end{minipage}
  \begin{minipage}[b]{\hsize}
    \centering
    \includegraphics[keepaspectratio, width=\hsize]{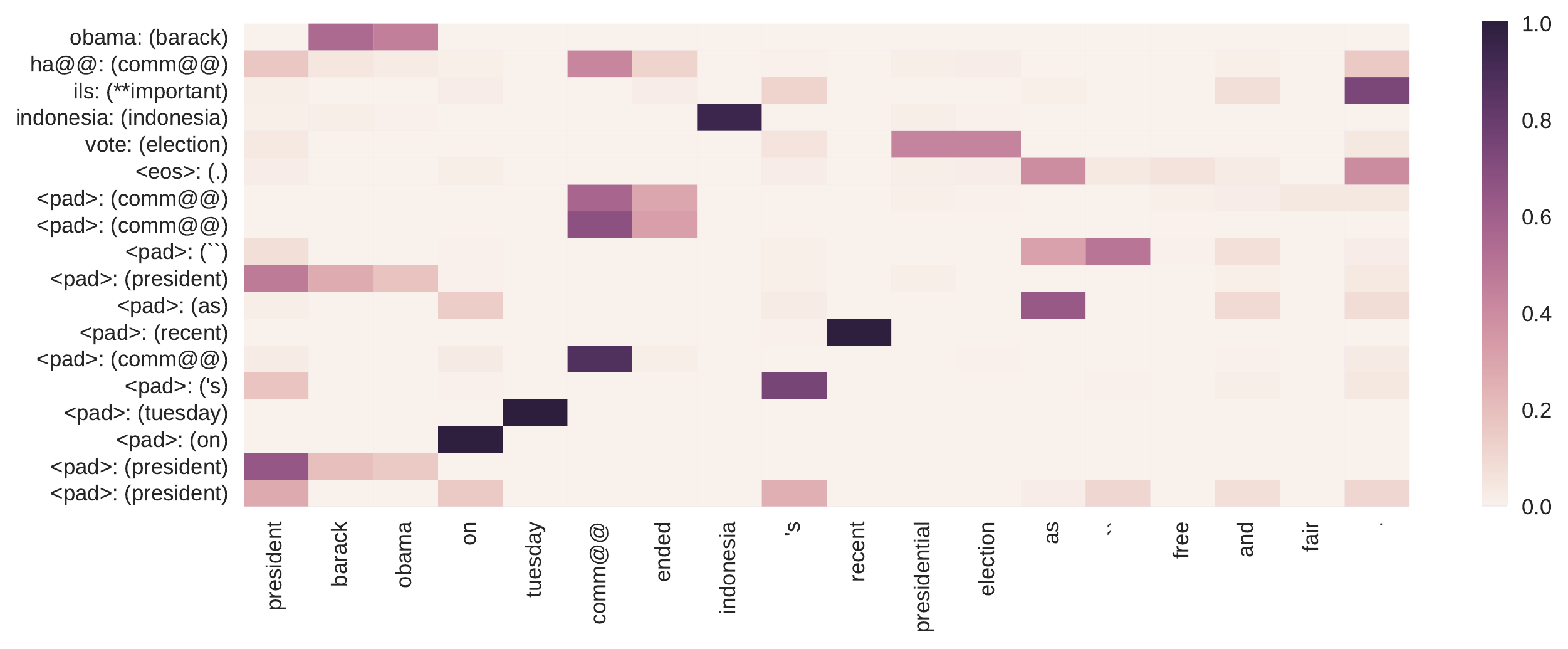}
    \subcaption{Source-side prediction of \spm}
  \end{minipage}
  \caption{
  SPM aligns the ``election'' with ``vote'', whereas \rencdec{} aligns ``vote'' with sentence period.
  }\label{fig:heatmap02}
\end{figure*}

\begin{figure*}[t]
  \begin{minipage}[b]{\hsize}
    \centering
    \includegraphics[keepaspectratio, width=\hsize]{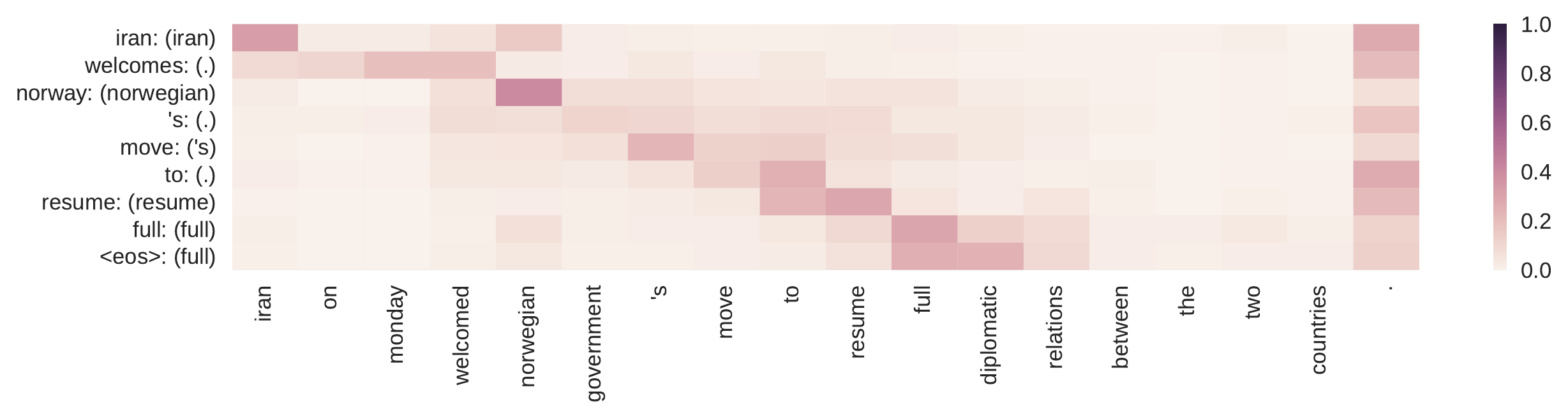}
    \subcaption{Attention distribution of \encdec}
  \end{minipage}
  \begin{minipage}[b]{\hsize}
    \centering
    \includegraphics[keepaspectratio, width=\hsize]{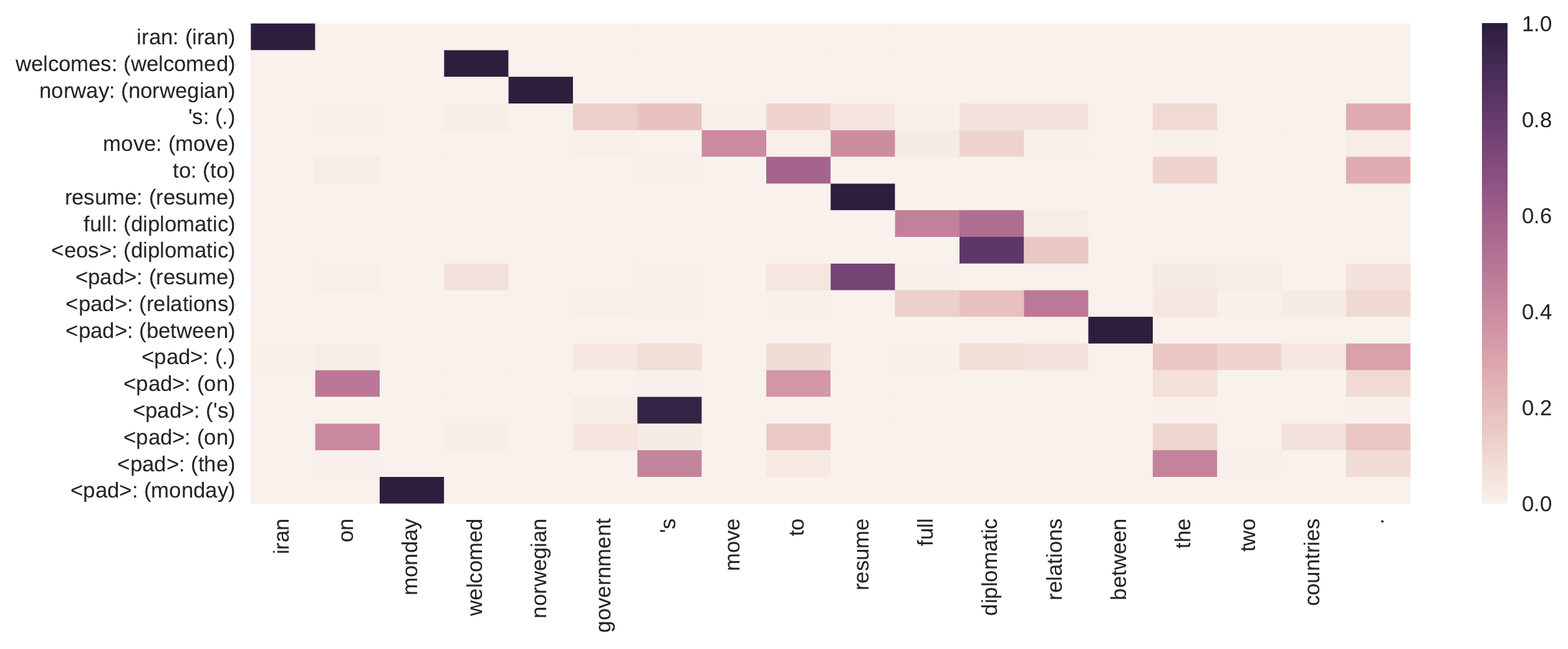}
    \subcaption{Source-side prediction of \spm}
  \end{minipage}
  \caption{
  The SPM aligns ``welcomes'' with ``welcomed.''
  On the other hand, \rencdec{} aligns ``welcomes'' with the sentence period.
  }\label{fig:heatmap03}
  \end{figure*}

Figures \ref{fig:heatmap01}, \ref{fig:heatmap02} and \ref{fig:heatmap03} are the extra visualizations of SPM and attention.
We created each figure with the procedure described in Section \ref{subsec:visualize}.

\section{Obtained Alignments}
We analyzed the source-side prediction to investigate the alignment that the SPM acquires.
We randomly sampled 500 source-target pairs from Gigaword Test (Ours), and fed them to \rspm{}.
For each decoding time step $j$, we created the alignment pair by comparing the target-side token $\bm{y}_{j}$ with the token with the highest probability over the source-side probability distribution $\bm{q}_{j}$.
Table~\ref{tb:alignment} summarizes the examples of the obtained alignments.
The table shows that the SPM aligns various type of word pair, such as the verb inflection and paraphrasing to the shorter form.

\begin{table*}[]
\centering
\small
\begin{tabularx}{\textwidth}{cX}
\toprule
\textbf{Type}                         & \multicolumn{1}{c}{Aligned Pairs: (Target-side Token, SPM Prediction)} \\
\midrule
Verb Inflection              & (calls, called), (release, released), (win, won), (condemns, condemned), (rejects, rejected), (warns, warned)           \\
\midrule
Paraphrasing to Shorter Form & (rules, agreement), (ends, closed), (keep, continued), (sell, issue), (quake, earthquake), (eu, european)             \\
\midrule
Others                       & (tourists, people), (dead, killed), (dead, died), (administration, bush), (aircraft, planes), (militants, group) \\
\bottomrule
\end{tabularx}
\caption{Examples of the alignment that the SPM acquired}
\label{tb:alignment}
\end{table*}

\end{document}